\documentclass[10pt,twocolumn,letterpaper]{article}
\usepackage{xr}
\usepackage[final]{iccv}      %

\usepackage{siunitx}
\newcommand{\secref}[1]{Sec.~\ref{#1}}

\newcommand{\eqnref}[1]{Eq.~\eqref{#1}}

\makeatletter
\DeclareRobustCommand\onedot{\futurelet\@let@token\@onedot}
\def\@onedot{\ifx\@let@token.\else.\null\fi\xspace}
\def\eg{e.g\onedot} 
\def\ie{i.e\onedot} 
 
\def\etc{etc\onedot}

\makeatother

\newcommand{\boldparagraph}[1]{\vspace{3pt}\noindent{\bf #1}\xspace}

\definecolor{darkgreen}{rgb}{0,0.7,0}
\definecolor{darkblue}{RGB}{31,119,180}
\definecolor{darkred}{RGB}{214,39,40}

\newcommand{\panos}{
\begin{figure*}[t!]
\centering
  \includegraphics[width=\linewidth]{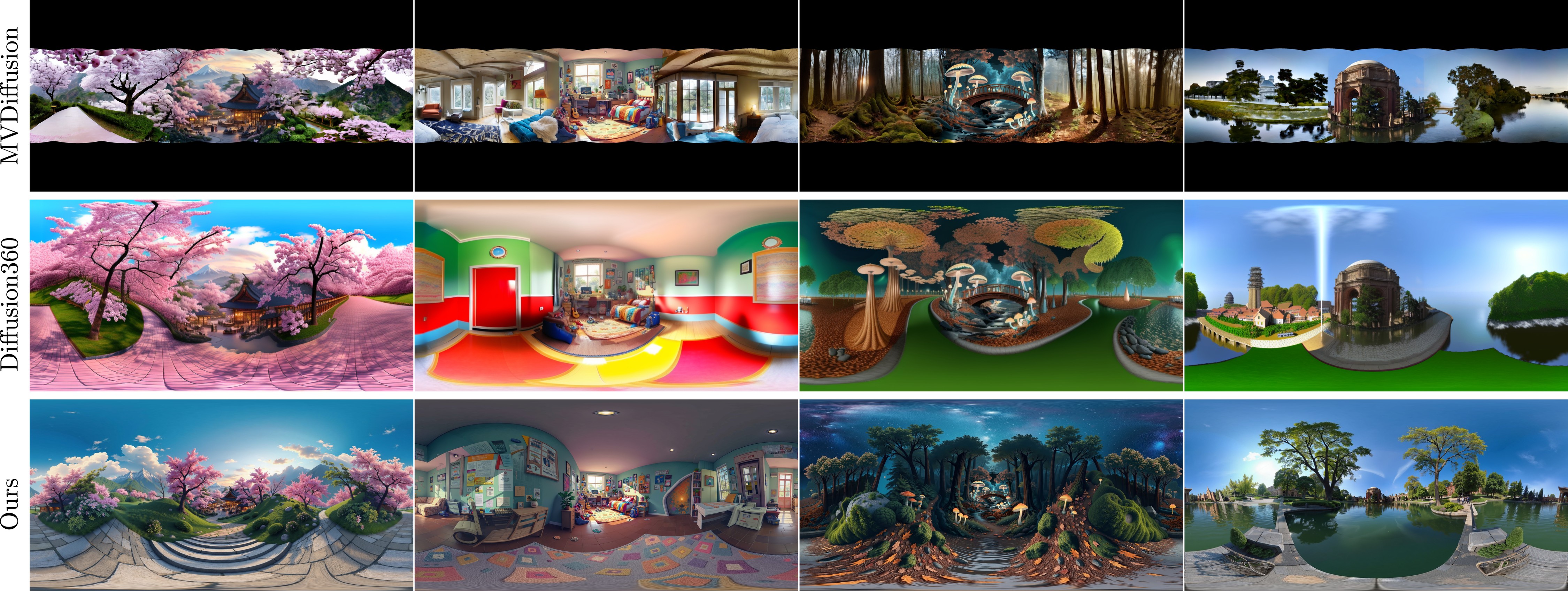}
  \caption{\textbf{Panorama Synthesis:} We show generated 360 panoramas from a single input image by our method. The reconstructions are consistent and result in accurate 3DGS scenes as visible in Fig.~\ref{fig:worlds}. }
  \label{fig:panos}
  \vspace{-5pt}
\end{figure*}
}

\newcommand{\worlds}{
\begin{figure*}[t!]
\centering
  \includegraphics[width=\linewidth]{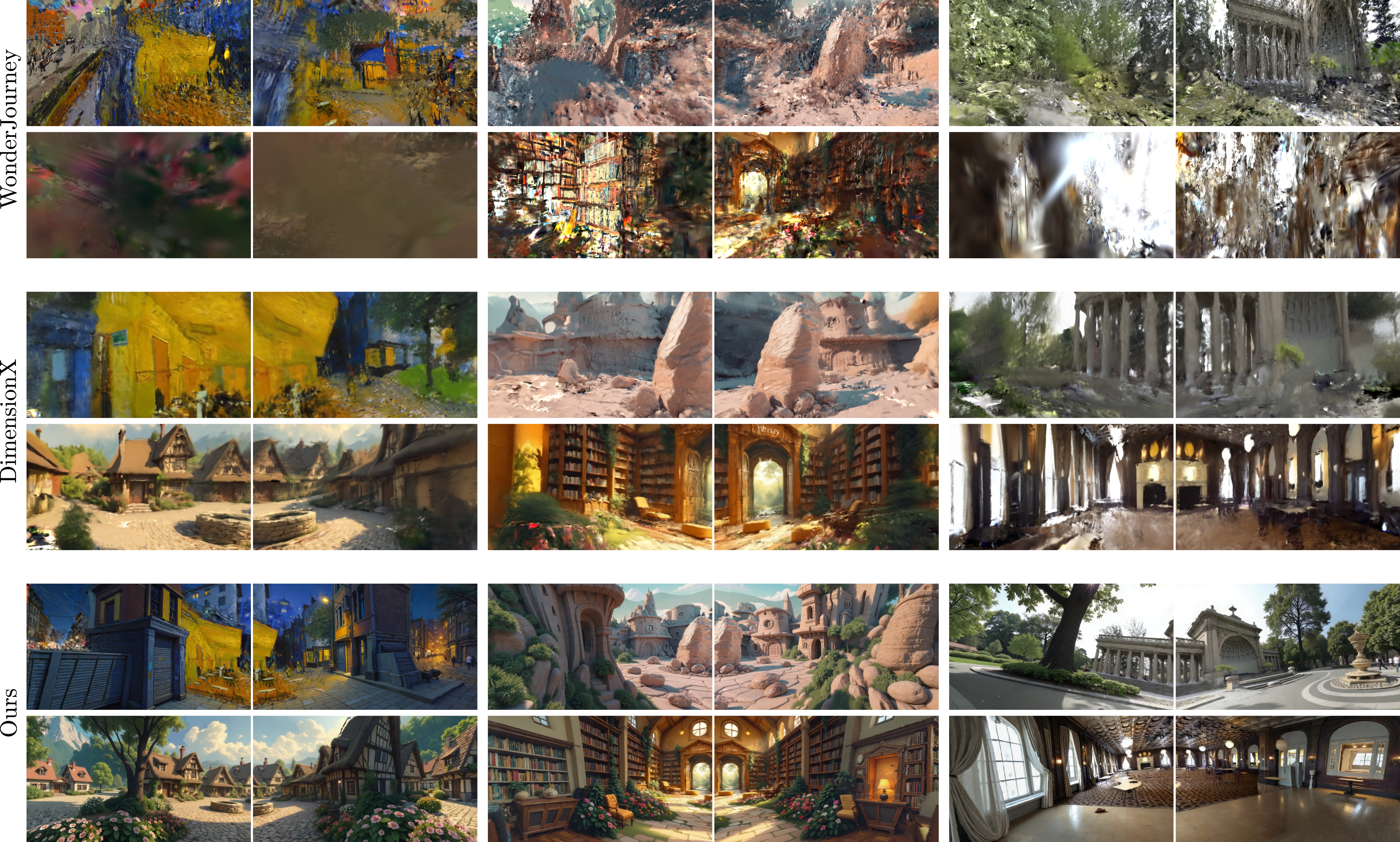}
  \caption{\textbf{3D Worlds:} Our method estimates 360-degree scenes given only a single input image. The proposed method clearly outperforms other baselines such as DimensionX~\cite{sun2024dimensionx} and WonderJourney~\cite{yu2023wonderjourney}, both qualitatively and quantitatively (Table~\ref{tab:gseval}). These baselines struggle to generate consistent 3D scenes. }
  \label{fig:worlds}
  \vspace{-5pt}
\end{figure*}
}

\newcommand{\worldsOurs}{
\begin{figure*}[t!]
\centering
  \includegraphics[width=\linewidth]{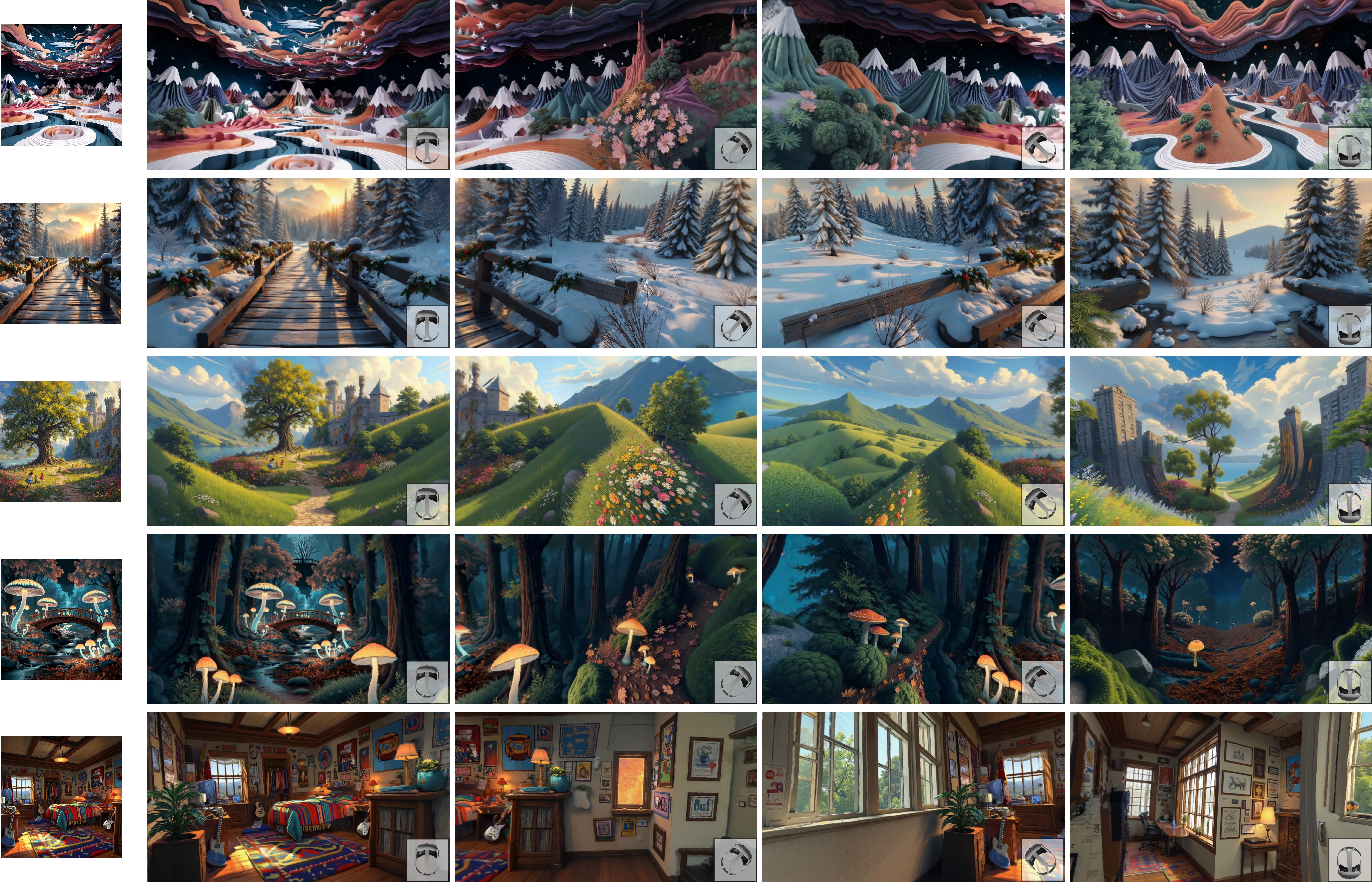}
  \caption{\textbf{3D Worlds:} Images rendered from the 3DGS representation generated by our pipeline, given only the single image shown on the left. The orientation of the VR headset in the bottom right corner highlights the direction of the novel views.} 
  \label{fig:worldsOurs}
  \vspace{-15pt}
\end{figure*}
}

\newcommand{\distortion}{
    \begin{figure}[h!]
      \centering
      \begin{subfigure}{0.49\linewidth}
            \includegraphics[width=\linewidth]{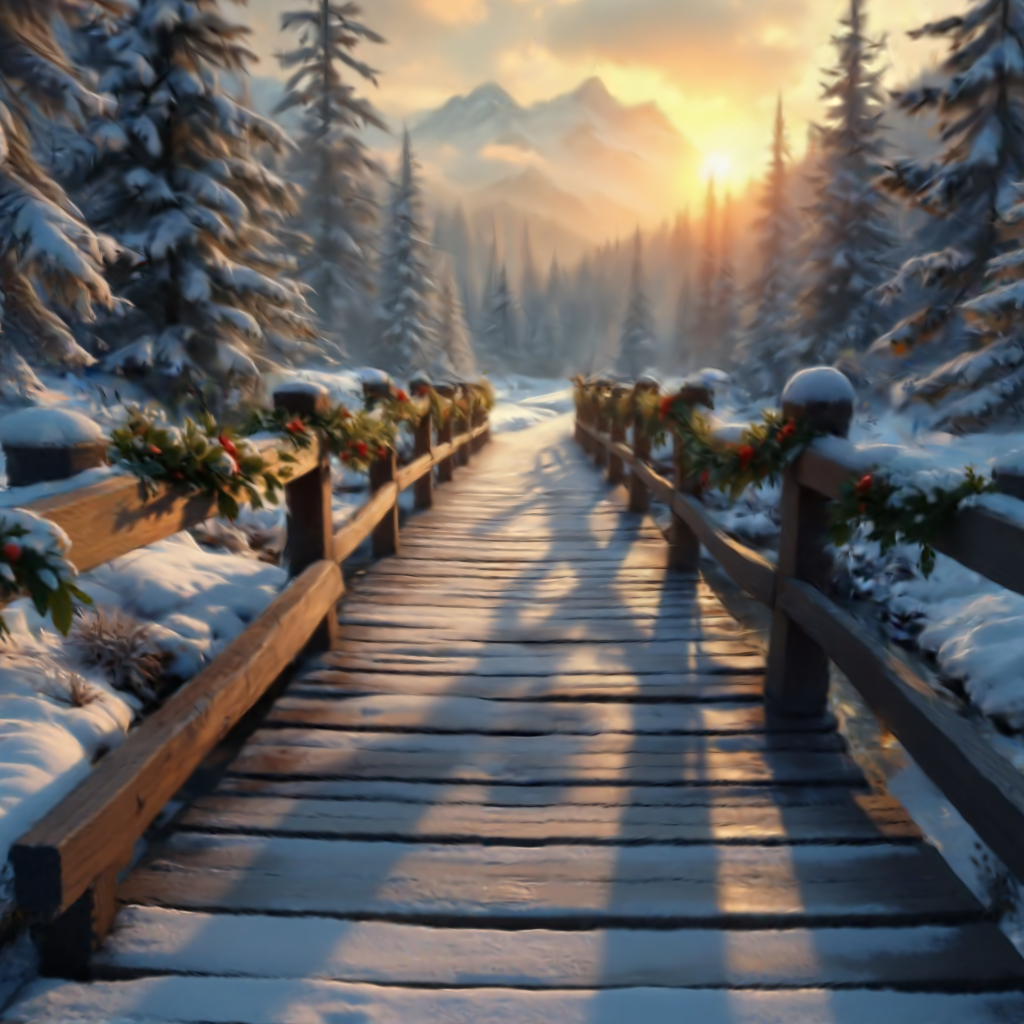}
        \caption{w/o distortion}
      \end{subfigure}
      \hfill
      \begin{subfigure}{0.49\linewidth}
            \includegraphics[width=\linewidth]{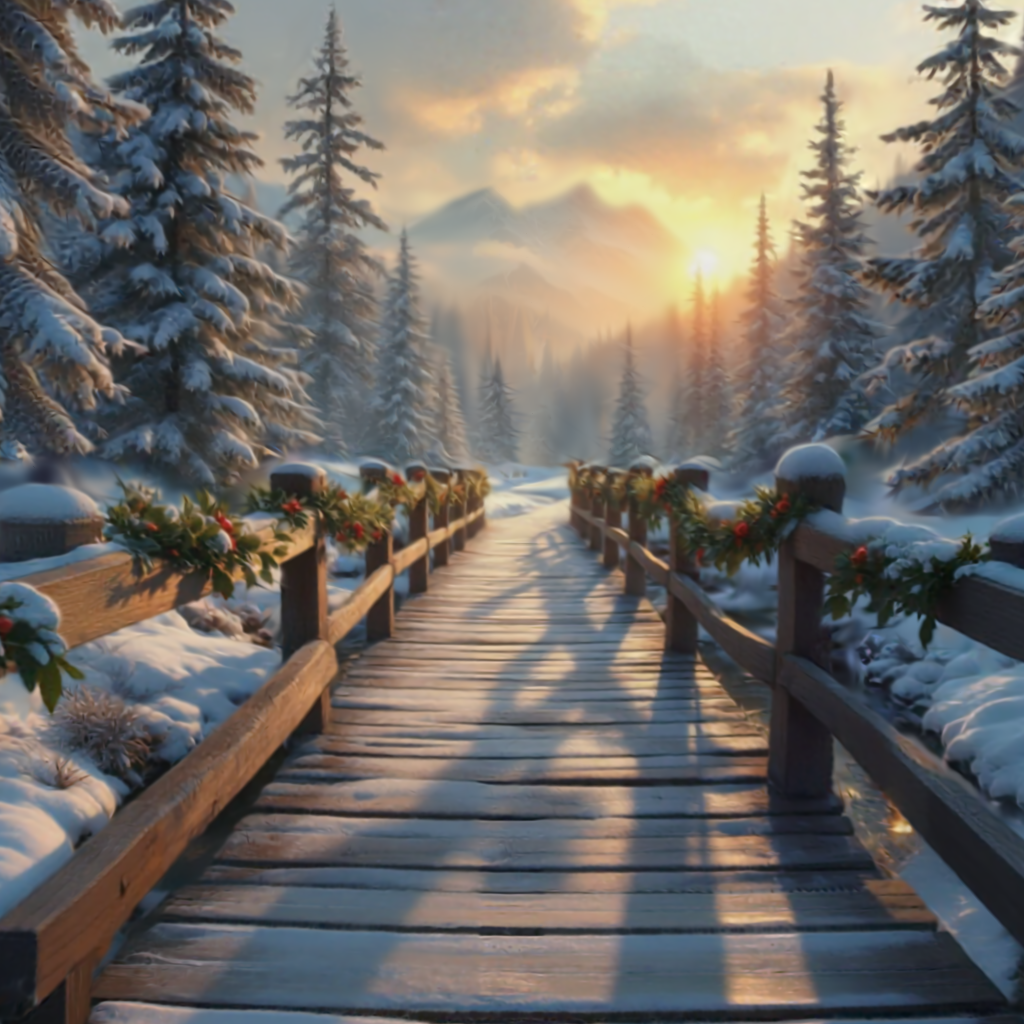}
        \caption{with distortion}
      \end{subfigure}
      \caption{\textbf{Trainable Image Distortion: }Integrating a trainable image distortion results in sharper reconstructions, compare \eg the details on the tree branches.}
      \label{fig:distortion}
      \vspace{-15pt}
    \end{figure}
}

\newcommand{\methodLift}{
    \begin{figure}
      \centering
      \begin{subfigure}{0.49\linewidth}
            \includegraphics[width=\linewidth]{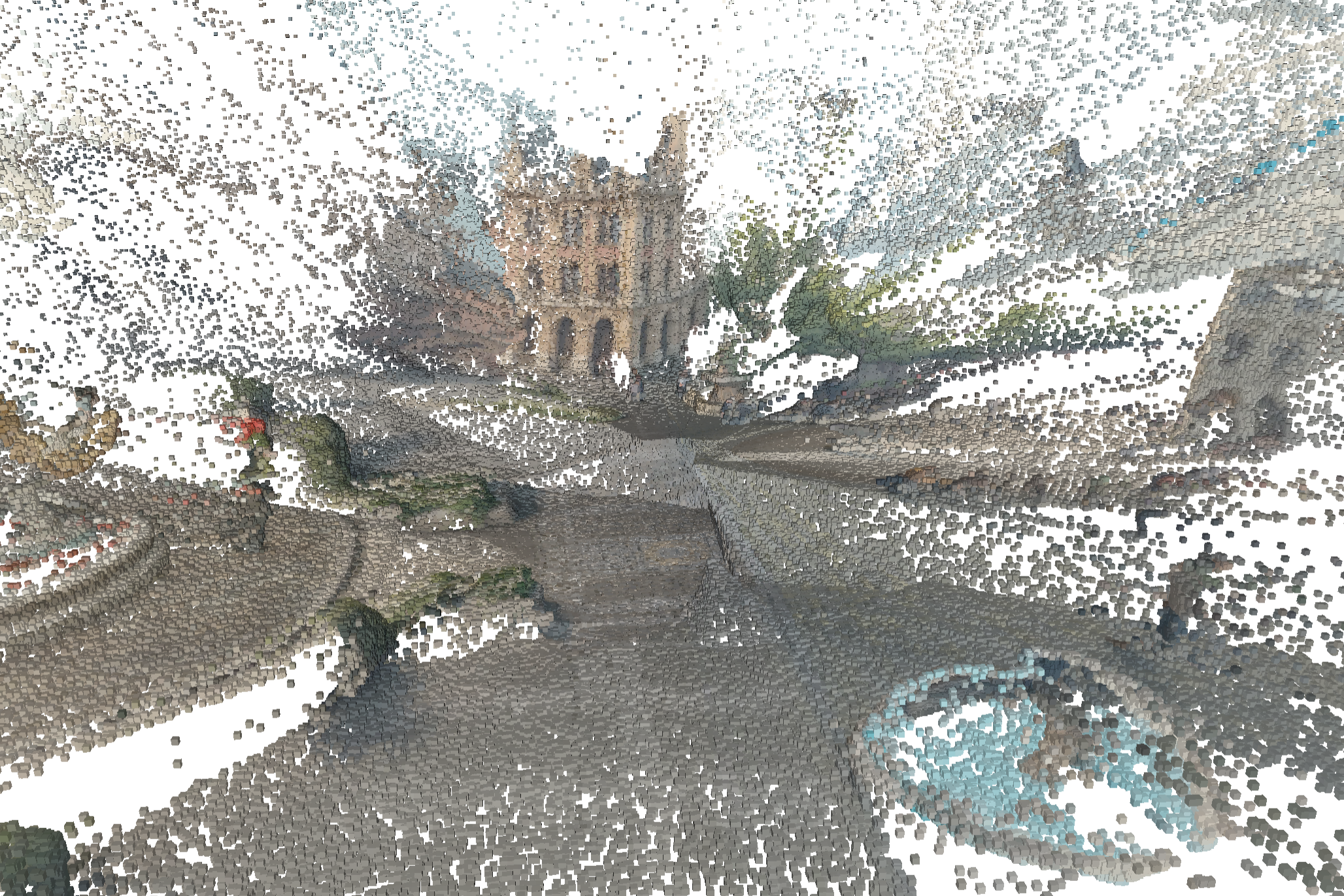}
        \caption{Metric3Dv2}
        \label{fig:method-metric}
      \end{subfigure}
      \hfill
      \begin{subfigure}{0.49\linewidth}
            \includegraphics[width=\linewidth]{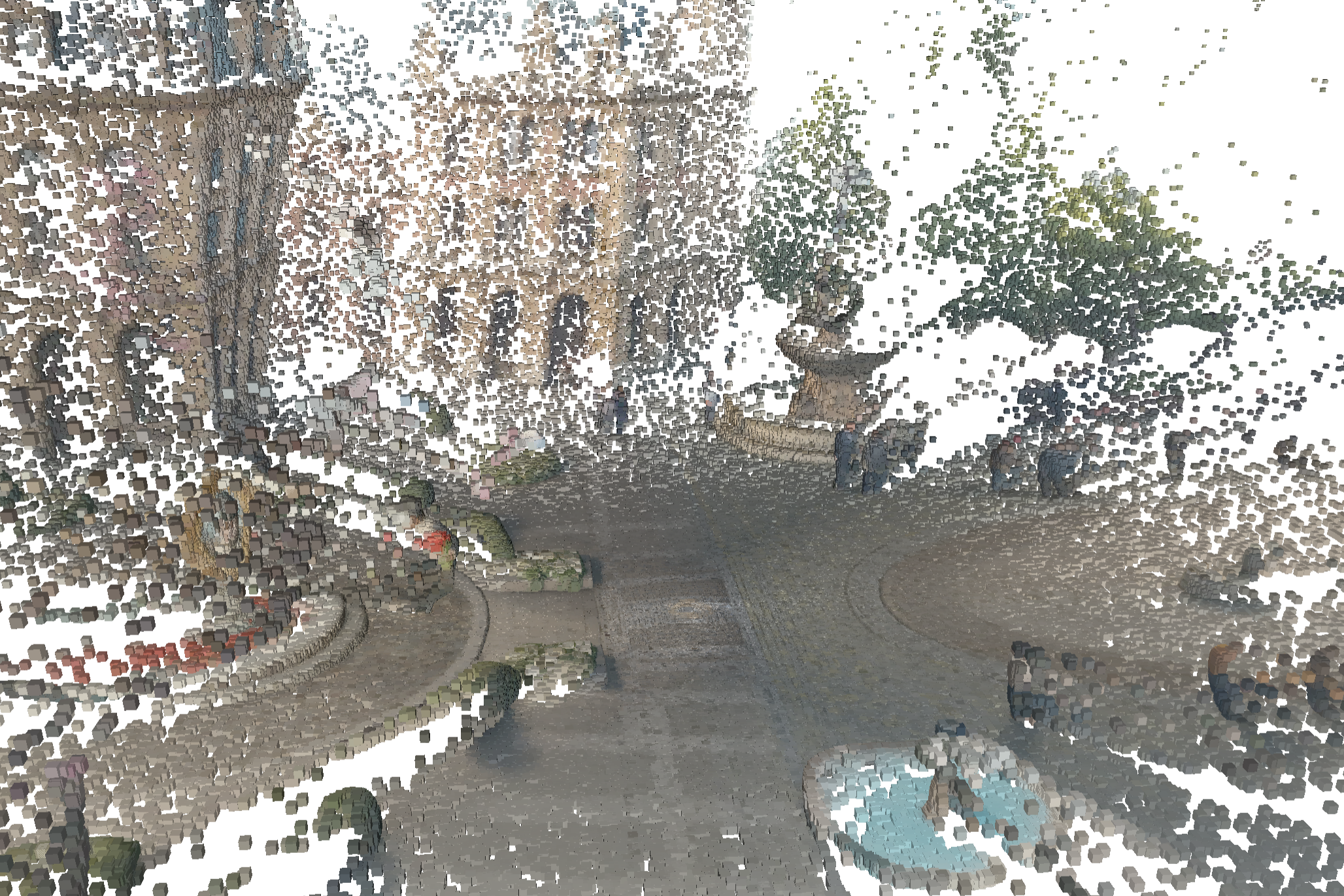}
        \caption{MoGE}
        \label{fig:method-moge}
      \end{subfigure}
      \vspace{-5pt}
      \caption{\textbf{Panorama Lifting: }Comparison of the lifted point clouds using metric depth estimation (Metric3Dv2) and monocular depth estimation (MoGE). The metric point cloud is distorted and contains prominent artifacts around the center.}
      \label{fig:methodLift}
      \vspace{-15pt}
    \end{figure}
}

\newcommand{\appAblationsA}{
\begin{figure*}[t!]
\centering
  \includegraphics[width=\linewidth]{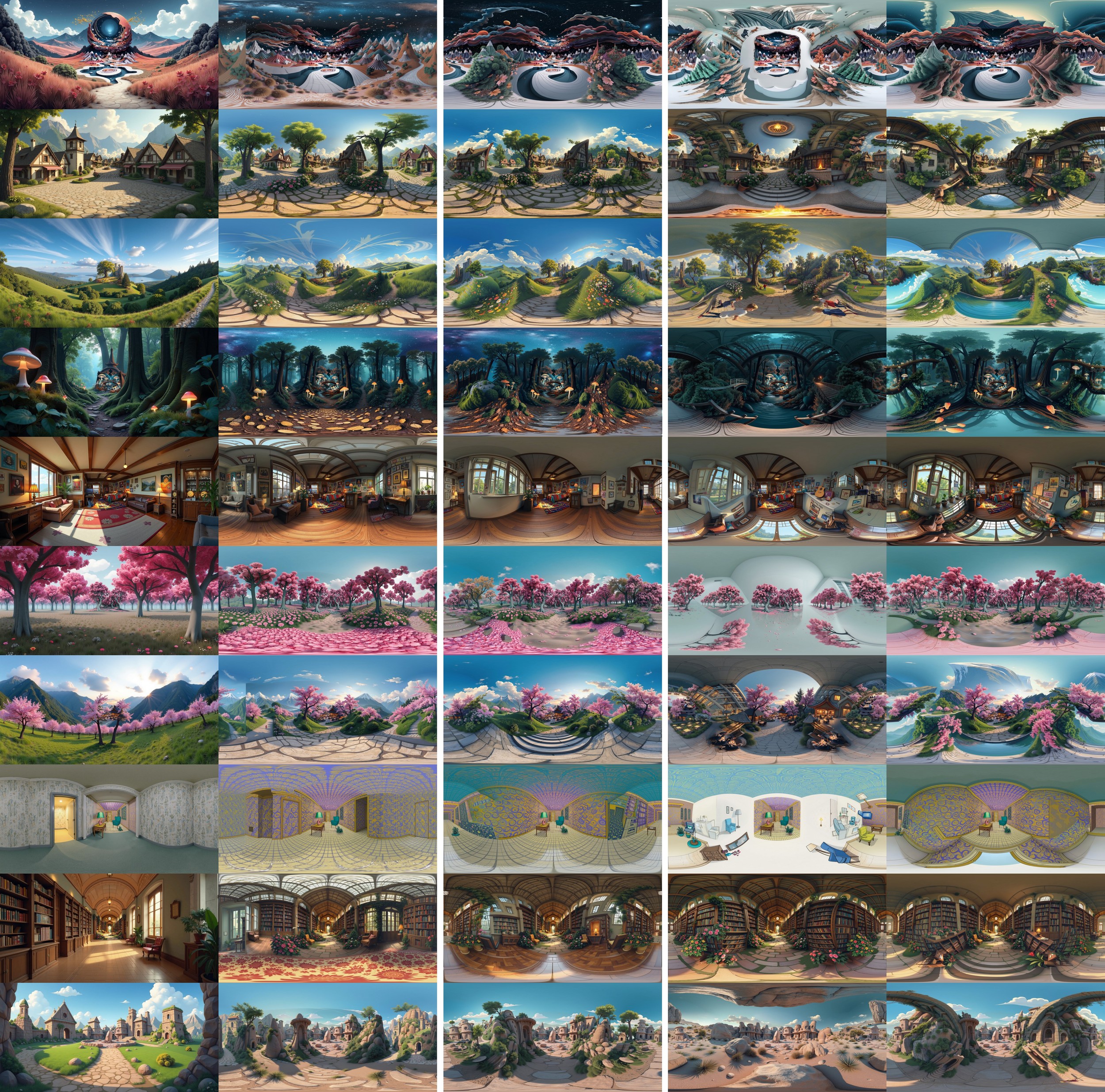}
  \caption{\textbf{Qualitative Ablation For Panorama Synthesis: }We compare different heuristics for progressive panorama synthesis and prompt generation. From left to right: Ad-Hoc, Sequential, Anchored, Prompt is caption generated from the input image, Non-specific prompt from vision-language model.}
  \label{fig:appAblationsA}
\end{figure*}
}

\newcommand{\appAblationsB}{
\begin{figure*}[t!]
\centering
  \includegraphics[width=\linewidth]{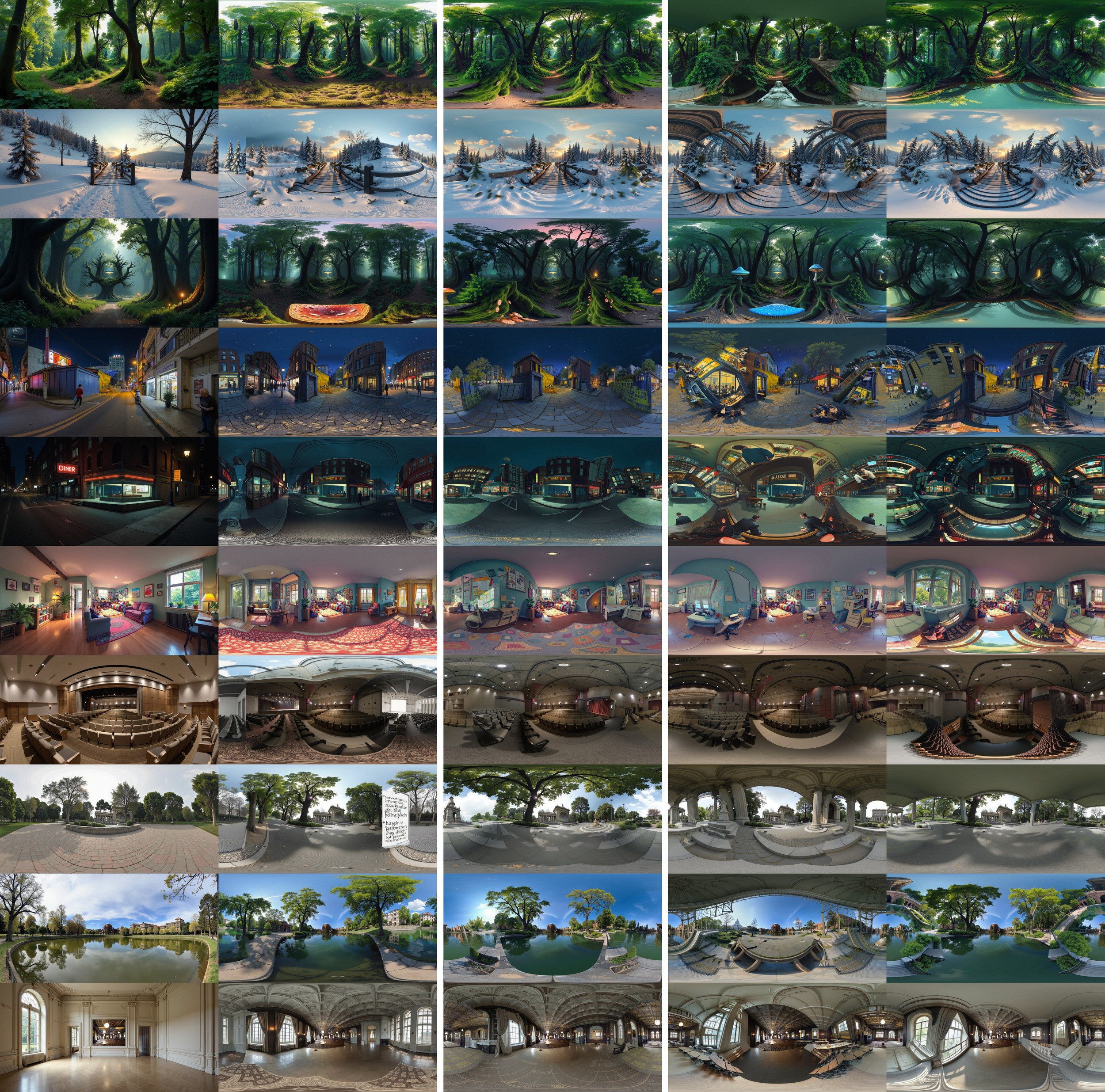}
  \caption{\textbf{Qualitative Ablation For Panorama Synthesis: }We compare different heuristics for progressive panorama synthesis and prompt generation. From left to right: Ad-Hoc, Sequential, Anchored, Prompt is caption generated from the input image, Non-specific prompt from vision-language model.}
  \label{fig:appAblationsB}
\end{figure*}
}

\newcommand{\fail}{
\begin{figure}[t!]
\centering
  \includegraphics[width=\linewidth]{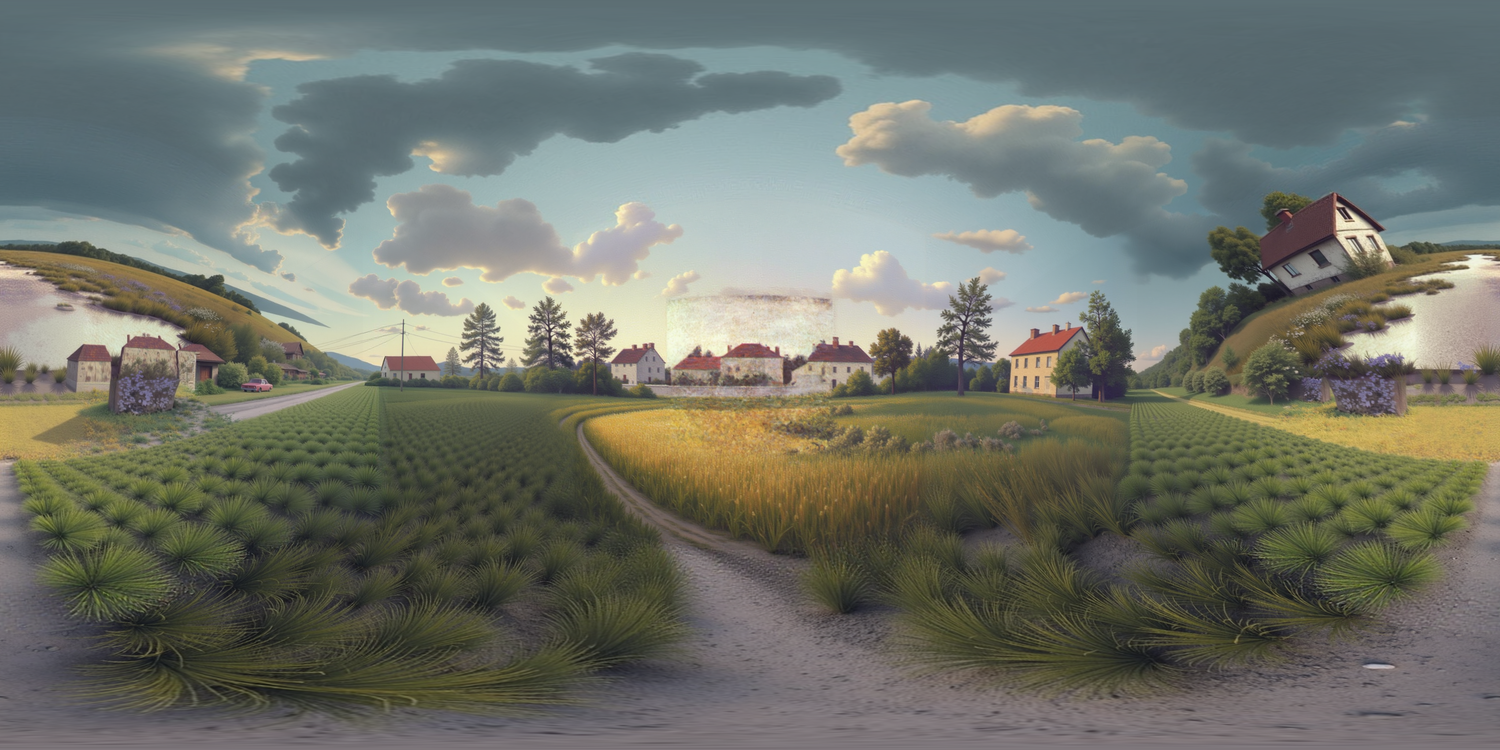}
  \caption{\textbf{Failure Case For Panorama Synthesis: } For challenging input images, \eg artworks with a distinct style, the inpainting model can struggle to adapt the global style of the panorama, resulting in visible border artifacts around the input image. Further, even with our anchored heuristic, the spatial layout of the panorama can sometimes be imperfect.}
  \label{fig:fail}
\end{figure}
}

\newcommand{\method}{
    \begin{figure*}
      \centering
      \begin{subfigure}{0.594\linewidth}
            \includegraphics[width=\linewidth]{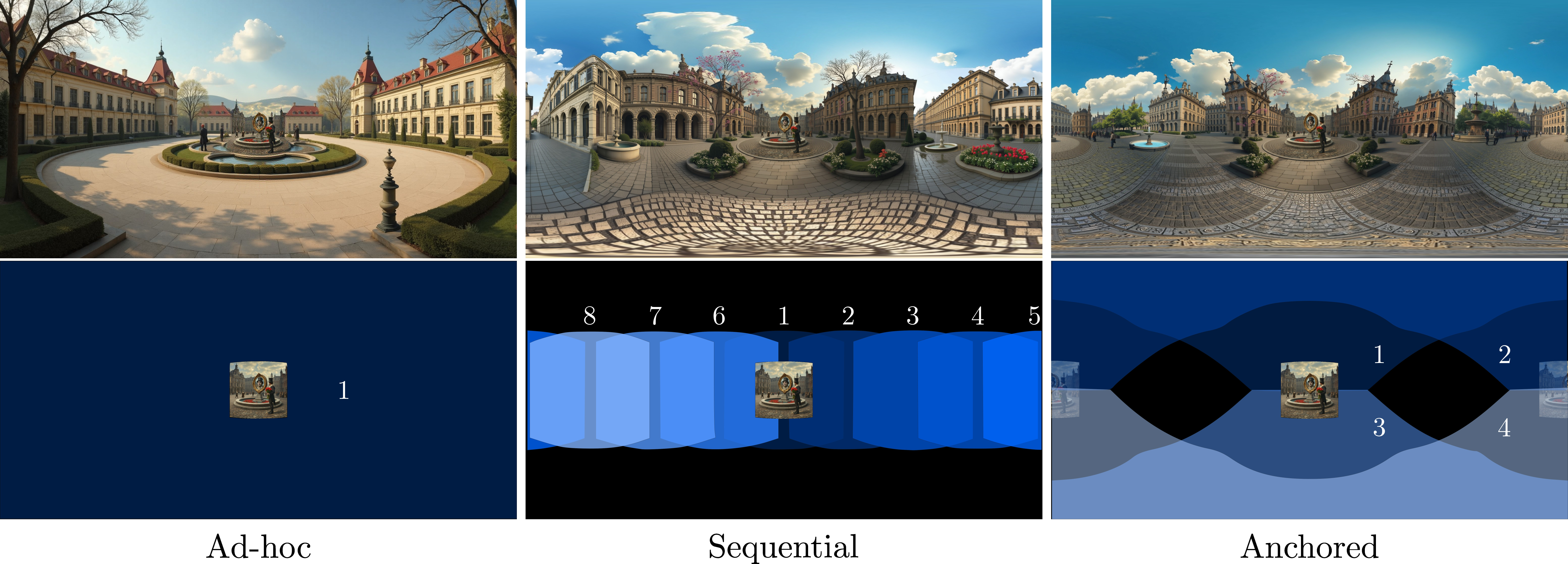}
        \caption{\textbf{Progressive Panorama Synthesis.} The white numbers indicate the order of the generated views. To avoid clutter, we only highlight the first generated views.
        \textbf{Ad-hoc}: The model is asked to directly outpaint the panorama image in a single step. \textbf{Sequential}: The camera rotates right, then left before inpainting the sky and ground. \textbf{Anchored}: The input image is duplicated to the backside to anchor sky and ground synthesis which are generated first.}
        \label{fig:method-panos}
      \end{subfigure}
      \hfill
      \begin{subfigure}{0.396\linewidth}
            \includegraphics[width=\linewidth]{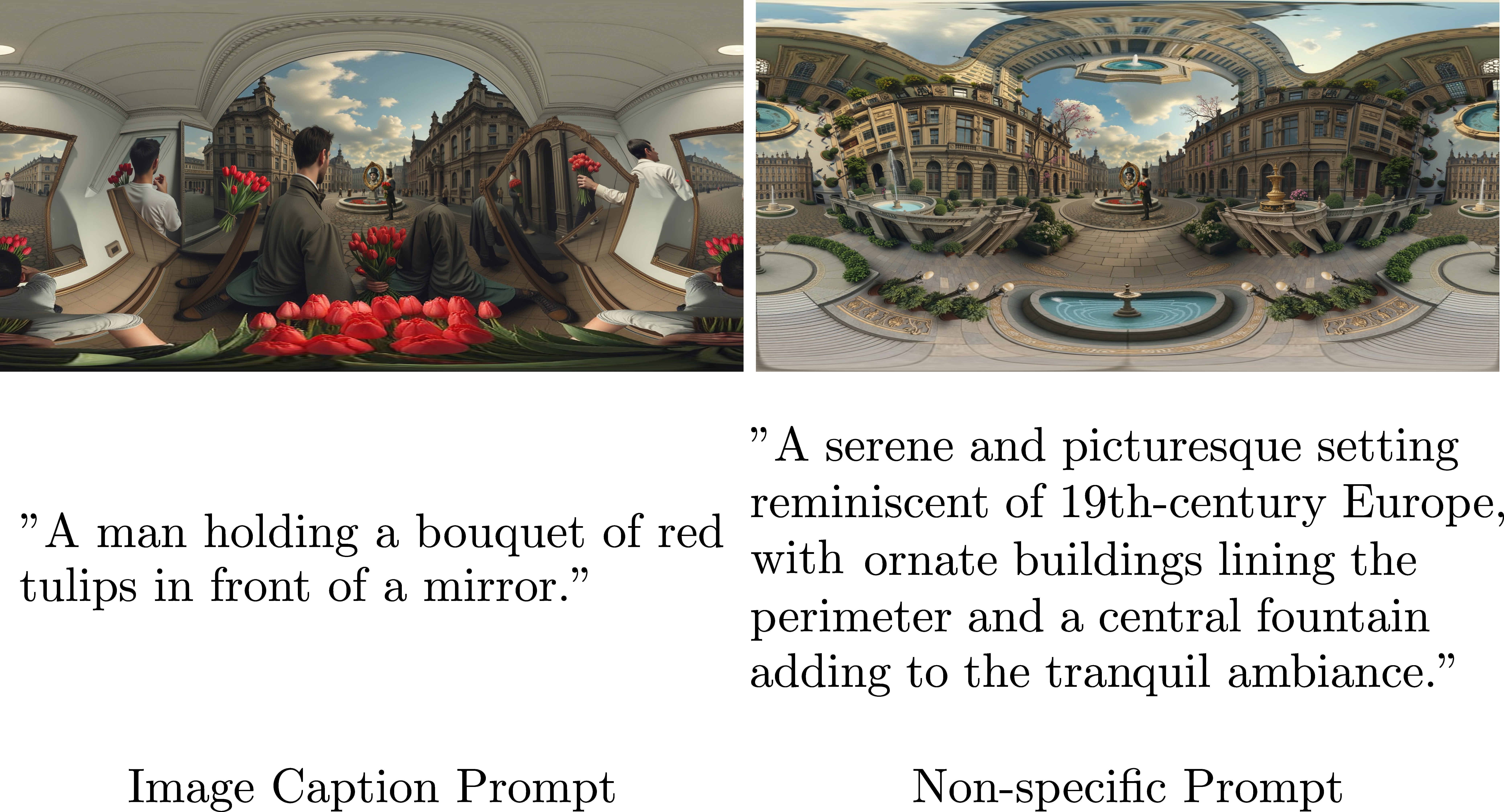}
        \caption{\textbf{Prompt Generation.} Left: The panorama prompt is a caption generated from the input image. Right: Using a non-specific prompt for the full panorama. For comparison, \cref{fig:method-panos} (Anchored), shows the generated panorama using the non-specific prompt with individual prompts for sky and ground.}
        \label{fig:method-prompts}
      \end{subfigure}
          \vspace{-5pt}
      \caption{\textbf{Panorama Synthesis: }Generated panorama images (top) and the respective synthesis heuristic (bottom).}
      \label{fig:method}
  \vspace{-5pt}
    \end{figure*}
}

\newcolumntype{H}{>{\setbox0=\hbox\bgroup}c<{\egroup}@{}}       %

\newcommand{\panorama}{
\begin{table*}[t]
    \centering
    \setlength{\tabcolsep}{12pt}
    \resizebox{\linewidth}{!}{
    \begin{tabular}{lS[round-precision=1, table-format=2.1]S[round-precision=1, table-format=1.1]S[round-precision=1, table-format=2.1]S[round-precision=1, table-format=1.1]S[round-precision=1, table-format=2.1]S[round-precision=1, table-format=2.1]S[round-precision=1, table-format=1.1]S[round-precision=1, table-format=1.1]}
    \toprule
                 & \multicolumn{4}{c}{WorldLabs Input Images}                                      & \multicolumn{4}{c}{Tanks and Temples Advanced}                                  \\
                 \cmidrule(lr){2-5}
                 \cmidrule(lr){6-9}
                 & {BRISQUE$\downarrow$} & {NIQE$\downarrow$} & {Q-Align$\uparrow$} & {CLIP-I$\uparrow$} & {BRISQUE$\downarrow$} & {NIQE$\downarrow$} & {Q-Align$\uparrow$} & {CLIP-I$\uparrow$} \\
                 \midrule
    MVDiffusion  & 51.5208740234375                &  6.769153575229901          &      2.8925193504050926               &     79.43351171634815             & 52.61330785506811                 & 6.713683278402807               &        2.863994891826923             &        78.34053611755371          \\
    Diffusion360 &      81.89372875072338               &      11.684867320125973            &         1.9109474464699074            &             75.10432917983444                   &  82.39946570763222          &    11.42256556713366                   &      2.0400766225961537               &        74.51147563640887           \\
    Ours         &       \bfseries 36.33177580656829             &   \bfseries   6.0078275691510195            &       \bfseries 3.484763816550926              &        \bfseries  81.88413049556591        &    \bfseries    36.64373779296875             &    \bfseries   5.905131539508223           &    \bfseries 3.3398030598958335                 &                 \bfseries 81.72967354456584 \\
    \bottomrule
    \end{tabular}
    }
    \caption{\textbf{Panorama Synthesis: } We assess image quality (BRISQUE, NIQE, Q-Align) and the alignment with the input image (CLIP) for panorama images at resolution 2048$\times$4096 pixels.}
    \label{tab:panorama}
    \vspace{-10pt}
\end{table*}
}

\newcommand{\inpainting}{
\begin{table}[t]
    \centering
    \setlength{\tabcolsep}{4pt}
    \resizebox{\linewidth}{!}{
    \begin{tabular}{lS[round-precision=1, table-format=2.1]S[round-precision=1, table-format=1.1]S[round-precision=1, table-format=2.1]S[round-precision=1, table-format=1.1]}
    \toprule
                                   & \multicolumn{4}{c}{ScanNet++}                                                    \\ \cmidrule(lr){2-5}
                                   & {BRISQUE$\downarrow$} & {NIQE$\downarrow$} & {Q-Align$\uparrow$} & {PSNR$\uparrow$} \\
        \midrule
    ControlNet, fwd warp         &    50.17789330812964                  &    6.520904687390587              &        3.454053217821782             &      11.983145567450192           \\
    ControlNet, fwd-bwd warp &        46.16563793220142             &    6.485292353622596             &      3.4901376856435644             &       15.879666640026734         \\
    \bottomrule
    \end{tabular}
    }
    \caption{\textbf{Point Cloud-Conditioned Inpainting: } We assess image quality (BRISQUE, NIQE, Q-Align) and the alignment with the input image (MSE) for inpainted images at resolution 576$\times$1024.}
    \label{tab:pcdInpainting}
    \vspace{-15pt}
\end{table}
}

\newcommand{\gseval}{
\begin{table*}[t]
    \centering
    \setlength{\tabcolsep}{12pt}
    \resizebox{\linewidth}{!}{
    \begin{tabular}{lS[round-precision=1, table-format=2.1]S[round-precision=1, table-format=1.1]S[round-precision=1, table-format=2.1]S[round-precision=1, table-format=2.1]S[round-precision=1, table-format=2.1]S[round-precision=1, table-format=1.1]}
    \toprule
                  & \multicolumn{3}{c}{WorldLabs Input Images}                   & \multicolumn{3}{c}{Tanks and Temples Advanced}               \\
                  \cmidrule(lr){2-4}
                  \cmidrule(lr){5-7}
                  & {BRISQUE$\downarrow$} & {NIQE$\downarrow$} & {Q-Align$\uparrow$} & {BRISQUE$\downarrow$} & {NIQE$\downarrow$} & {Q-Align$\uparrow$} \\
                  \midrule
    WonderJourney &           50.97386491602922          &         5.891806831168507         &          1.9062409577546295           &         45.05038324991862            &      5.290653296891749            &        2.0011266072591147             \\
    DimensionX    &        64.79640139270742             &     7.839026838152951             &           1.7231547037760417          &         63.115580240885414            &      7.55069997780997            &        1.7072347005208333             \\ \midrule
    Ours + ViewCrafter         &          43.5431224681713           &        6.02098010922363       &      3.4233907063802085             &        42.88787714640299           &        5.787246496396279        &       3.2509765625    \\ 
    Ours + ControlNet          &        41.093436347113716           &        5.594889200829639          &        3.5147727683738426            &       39.5131721496582              &          5.272140338827723         &      3.397857666015625       \\       
    Ours + ControlNet + Refined GS         &      \bfseries  33.850771303530095          &   \bfseries  4.627499263738299         &   \bfseries  3.61832117151331    &  \bfseries   33.88949966430664          &      \bfseries   4.496837995600008           &    \bfseries    3.453119913736979  \\
    \bottomrule
    \end{tabular}
    }
    \caption{\textbf{Quality in VR: } We assess image quality of images rendered from the 3DGS representation at resolution 1024$\times$1024 pixels, using a field of view of 60 degrees.}
    \label{tab:gseval}
    \vspace{-10pt}
\end{table*}
}

\definecolor{iccvblue}{rgb}{0.21,0.49,0.74}
\usepackage[pagebackref,breaklinks,colorlinks,allcolors=iccvblue]{hyperref}
\def\paperID{6393} %
\def\confName{ICCV}
\def\confYear{2025}

\title{\vspace{-30pt}A Recipe for Generating 3D Worlds From a Single Image}

\author{Katja Schwarz\quad Denys Rozumnyi\quad Samuel Rota Bul\`o\quad Lorenzo Porzi\quad Peter Kontschieder\vspace{1em}\\
Meta Reality Labs Zurich, Switzerland
}

\begin{document}
\twocolumn[{
        \renewcommand\twocolumn[1][]{#1}%
    \maketitle
    \begin{center}
    \includegraphics[width=\linewidth, trim={0 0.0cm 0.0cm 0cm},clip]{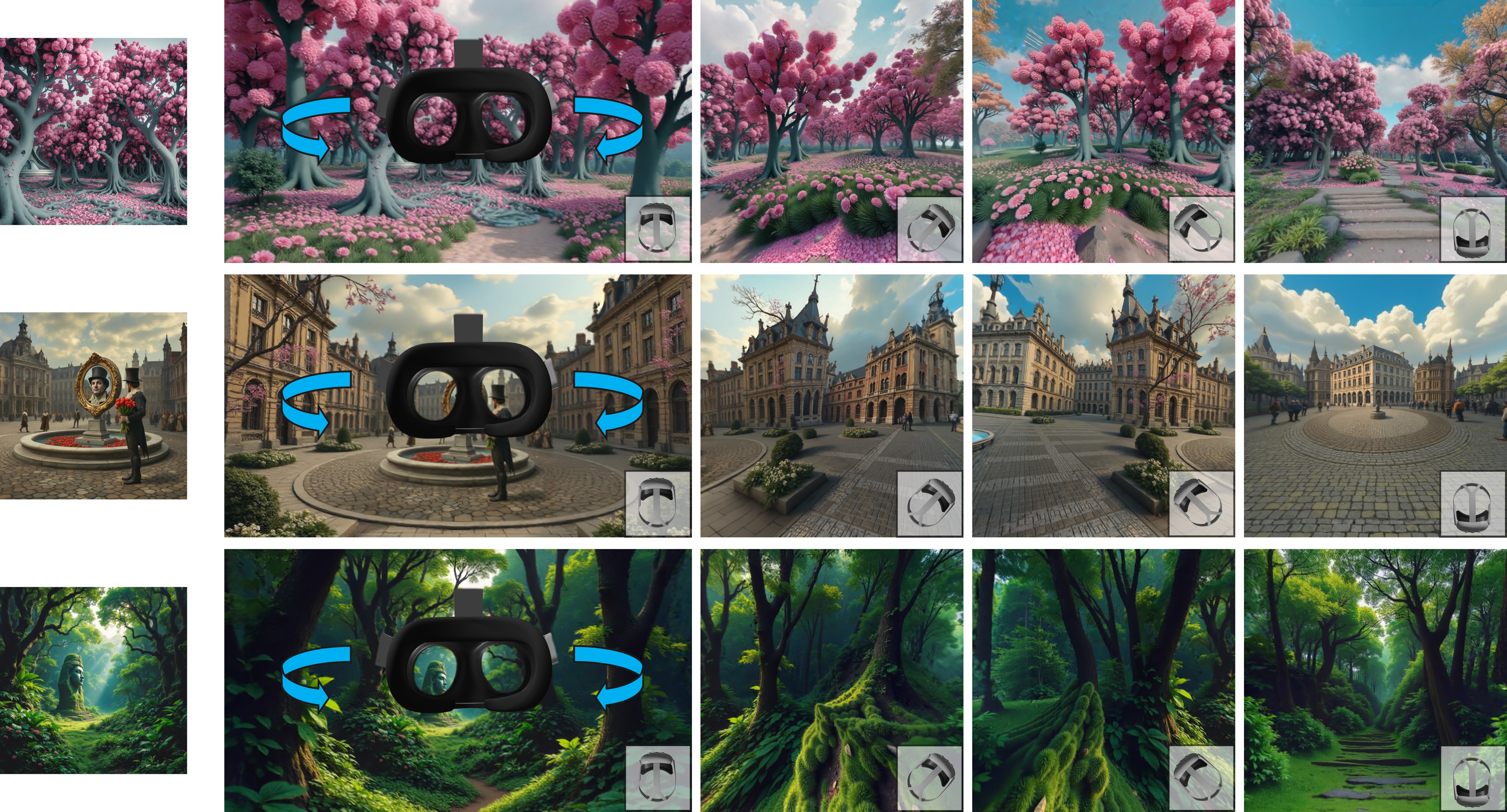}
         \captionof{figure}{\textbf{Overview: }Given a single input image, our pipeline generates a 360 degree world. The scene is parameterized by Gaussian Splats and can be explored on a VR headset within a cube with 2m side length. Project Page: \url{https://katjaschwarz.github.io/worlds/}}
        \label{fig:teaser}
    \end{center}
}]
\begin{abstract}
We introduce a recipe for generating immersive 3D worlds from a single image by framing the task as an in-context learning problem for 2D inpainting models. This approach requires minimal training and uses existing generative models. Our process involves two steps: generating coherent panoramas using a pre-trained diffusion model and lifting these into 3D with a metric depth estimator. We then fill unobserved regions by conditioning the inpainting model on rendered point clouds, requiring minimal fine-tuning. Tested on both synthetic and real images, our method produces high-quality 3D environments suitable for VR display. By explicitly modeling the 3D structure of the generated environment from the start, our approach consistently outperforms state-of-the-art, video synthesis-based methods along multiple quantitative image quality metrics.
\end{abstract}

\section{Introduction}
Leveraging image-guided 3D scene synthesis has the potential to disrupt traditional 3D content creation workflows, enabling the rapid generation of high-fidelity, plausible 3D environments. With increasing consumer interest and an ever-growing ecosystem of VR devices, there is a strong need for simple and user-friendly approaches to the generation of 3D content, propelling new applications in gaming, social experience apps, the way we marvel at art, \etc.\\
Generating 3D environments for VR from a single input image is a highly ambiguous and complex task. The ill-posed nature of this problem arises from the fact that multiple possible 3D scenes can be projected onto the same 2D image. It is nontrivial to provide a solution that retains consistency in the generated style and overall coherence of the result. Also, the quality of the generated 3D geometry has a significant impact on the overall VR experience, as incorrect 3D structures lead to view-dependent inconsistencies that can easily break the sense of immersion.\\
Recent advances in image and video generation models have shown promising results for the synthesis of high-quality 2D content. However, these models typically lack 3D consistency, leading to blurry and incorrect scene artifacts in areas where their 2D prior is supposed to provide consistent completion information for 3D scenes. 
Although autoregressive outpainting can be applied to a certain extent for covering up an inherent lack of 3D consistency in generative models, it typically leads to noticeable 360$^\circ$ stitching artifacts, which are among the most unpleasant effects in the single-image conditioned generation scenario. \\
In our work, we propose a simple and yet effective approach for single image to 3D scene generation, with novel view synthesis in VR as the primary mode of consumption in mind. Our solution decomposes the overall generation task into a two-step approach: 2D panorama synthesis and lifting the generated scene into a refined, three-dimensional space. The resulting virtual environment is designed to be both viewable and navigable within a 2-meter cube when experienced through a VR headset. \\
We frame 2D panorama synthesis as an in-context learning task for existing inpainting models. By incorporating a vision-language model for prompt generation, our approach can generate high fidelity panorama images without requiring any additional training. 
To lift the generated panorama into a refined and approximately metric three-dimensional space, we first apply monocular, metric depth estimation on rendered images. This works sufficiently well for images rendered from the panorama, but usually leaves empty spots in previously occluded areas or at large depth discontinuities emerging when the camera views are shifted (\ie, underwent a translation). We identify this as another inpainting task, and demonstrate that the inpainting model can quickly adapt to this setting, when fine-tuned with appropriate masks derived from the rendered point clouds.   
Finally, after generating sufficiently many views we leverage Gaussian Splatting (3DGS)~\cite{kerbl3Dgaussians} as a 3D representation, which can be efficiently trained, and rendered in real time. To account for minor remaining (local) inconsistencies between the generated multi-view images, we augment 3DGS with a distortion correction mechanism, leading to overall sharper and more detailed results. \\
We provide a comprehensive experimental section with qualitative and quantitative results, comparing our proposed single image to 3D generation method against state-of-the-art methods like WonderJourney~\cite{yu2023wonderjourney} and DimensionX~\cite{sun2024dimensionx}. We demonstrate substantial improvements across all relevant metrics measuring the alignment to the input image's appearance and on image quality metrics, following~\cite{zhou2024dreamscene360}. We also provide detailed ablations for our panorama generation and point cloud-conditioned inpainting steps. 
To summarize, our contributions for improving single image to 3D world generation are as follows:
\begin{itemize}
    \item We decompose 3D scene synthesis into two easier subproblems: panorama synthesis, and point cloud-conditional inpainting, enabling the generation of 360 degree navigable environments from a single input image. 
    \item We propose a novel approach to panorama generation inspired by visual in-context learning, leading to more consistent sky and ground synthesis while enhancing overall image quality. 
    \item For point cloud-conditioned inpainting, we propose a simple, yet efficient forward-backward warping strategy for fine-tuning a ControlNet with minimal training effort.
    \item We augment Gaussian Splatting (3DGS) with a distortion correction mechanism to account for minor remaining inconsistencies between generated multi-view images, leading to overall sharper and more detailed results.
\end{itemize}

\section{Related Work}

\boldparagraph{2D Generative Models.}
Diffusion Models (DMs)~\cite{ho2020denoising,sohldickstein2015deep,song2020score} are highly effective generative models that achieve state-of-the-art performance in text- and image-guided synthesis~\cite{nichol2021improved,rombach2021highresolution,dhariwal2021diffusion,ho2022cascaded,podell2023sdxl,ho2022imagenvideo,singer2022make,luo2023videofusion,esser2023structure,Esser2024rectifiedFlow}. ControlNet~\cite{zhang2023controlnet}, LoRA~\cite{Hu2022lora}, and IP-Adapter~\cite{Hu2023ipadapter} are widely used to make existing generative backbones controllable. In this work, we leverage T2I DMs, adapted to inpainting tasks with a ControlNet.

\worldsOurs
\boldparagraph{Scene Generation From a Single Image.} 
One line of research approaches scene generation as a 2.5D panorama synthesis task~\cite{tang2023MVDiffusion,Feng2023Diffusion360,kalischek2025cubediff}. These approaches fine-tune 2D DMs and, more recently, also enable the simultaneous synthesis of panoramas and depth information~\cite{Paliwal2024PanoDreamer}. MVDiffusion~\cite{tang2023MVDiffusion} generates eight horizontally-rotated sub-views of a panoramic image in parallel, using a standard DM augmented with correspondence-aware attention.
Diffusion360~\cite{Feng2023Diffusion360} combines latents across multi-directional views, both at the denoising and VAE stages, in order to generate consistent 360$^\circ$ views that seamlessly blend together.
The first stage of our approach similarly generates a panorama image, given a single input image. While existing approaches fine-tune diffusion backbones, we propose a training-free method that frames panorama synthesis as an in-context zero-shot learning task for existing inpainting models. By incorporating global context during the panorama generation process, we achieve both improved style consistency and image quality without the need for training, as we show in Sec.~\ref{sec:panorama_generation}.
Another line of works directly synthesize navigable 3D environments. Generally, these works follow one of two high-level frameworks: i) 3D-guided image inpainting, or ii) 3D- and camera-guided video diffusion.
Most works based on framework (i)~\cite{Shriram2024realmdreamer,Liang2024luciddreamer,yu2023wonderjourney,Hollein2023texttoroom,SceneScape,wang2024vistadream,Pu2024pano2room,Yu2024wonderWorld,Asija2024PanoInpainting} adopt a very similar underlying pipeline, alternating depth prediction, image warping to a novel viewpoint, and in-painting of disoccluded regions.
While similar to the approach we propose to lift our generated 2D panorama to 3D, these methods are typically unable to produce a fully immersive scene, notably struggling with outpainting towards the opposite direction of the initial view, as we show in Sec.~\ref{sec:3d_worlds}.
Works based on framework (ii)~\cite{Yu2024ViewCrafter,sun2024dimensionx,Muller2024MultiDiff,liang2024wonderland,Wallingford2024360videos,Xu2024Camco,Sargent2024zeronvs} aim to re-purpose video diffusion models for 3D synthesis, or 3D-consistent video synthesis.
ViewCrafter~\cite{Yu2024ViewCrafter} and MultiDiff~\cite{Muller2024MultiDiff} progressively construct a point cloud-based representation of the scene, and use it as a conditioning signal for a video diffusion model.
DimensionX~\cite{sun2024dimensionx} uses a diffusion model to generate a video sequence given a single image and a camera path, then reconstructs the scene from this video with a combination of DUSt3R~\cite{Wang2024CVPRDuster} and uncertainty-aware Gaussian Splatting.
Even with the surprisingly strong, latent understanding of 3D geometry modern video generation models posses, we show in Sec.~\ref{sec:3d_worlds} that our approach is able to produce higher quality 3D scenes.
We argue that the key advantage of our method is to simplify the inherently hard problem of synthesizing arbitrary novel views in 3D, into the two, individually easier tasks of panorama generation and 3D lifting.
Notably, DreamScene360 also considers panorama synthesis and lifting as separate tasks for 3D scene synthesis. However, DreamScene360 is purely text-conditioned and cannot generate an 3D scene from a given input image.

\section{Method}
Our key insight is that the task of generating a 3D environment from a single image, which is inherently complex and ambiguous, can be decomposed into a series of more manageable sub-problems, each of which can be addressed with existing techniques.
In this section, we provide a step-by-step recipe that outlines these sub-problems and explains how existing approaches can be adapted to effectively address them.
We divide our approach into two main parts: 2D panorama synthesis and lifting the generated scene into three-dimensional space. 
The resulting virtual environment is designed to be both viewable and navigable within a 2-meter cube when experienced through a VR headset.

\subsection{Panorama Generation}
\method
Starting with a single input image, we introduce a progressive approach that frames panorama synthesis as a zero-shot learning task for a pre-trained inpainting model. We use a text-to-image (T2I) diffusion model that is conditioned on a masked input image using a ControlNet~\cite{zhang2023controlnet}.
First, the input image is embedded into an equirectangular image by calculating the perspective to equirectangular projection. Let $u$ and $v$ denote normalized pixel coordinates in range $[-1, 1]$. The coordinates are mapped to angles $\theta$ and $\phi$
\begin{align}
    \theta = u \times \frac{\text{fov}_x}{2}, \quad
    \phi = v \times \frac{\text{fov}_y}{2}
\end{align}
where $\text{fov}_x$ and $\text{fov}_y$ are horizontal and vertical field of view. 
The spherical coordinates, \ie $(\theta, \phi)$, are then converted to equirectangular coordinates:
\begin{align}
\tilde{x} = \left(\frac{\theta + \pi}{2\pi}\right) \times W, \quad
\tilde{y} = \left(\frac{\phi + \frac{\pi}{2}}{\pi}\right) \times H
\end{align}
where $W$ and $H$ are width and height of the equirectangular image. Typically, the aspect ratio is chosen as 2:1.  
In practice, we estimate $\text{fov}_x$ with Dust3R~\cite{Wang2024CVPRDuster} and derive $\text{fov}_y$ assuming equal focal length along the $x$ and $y$ axes.\\
Inspired by visual in-context learning~\cite{Bar2022NeurIPS}, we progressively outpaint this panorama image by rendering overlapping perspective images from it, see~\cref{fig:method-panos} for an illustration. 
We investigate three different heuristics for progressive synthesis: i)~\textbf{Ad-hoc}: We ask the model directly to synthesize a panorama image by appending "equirectangular image, panorama" to the prompt~(\cref{fig:method-panos}, left). While the generated image is reasonable, it does not have the correct equirectangular distortion for sky and ground.
ii)~\textbf{Sequential}: We rotate the camera 180 degree right and left, and then fill in sky and ground~(\cref{fig:method-panos}, middle). The middle of the panorama image is coherent but the ground does not match the scene. Since each image from the rotation is generated without global context, connecting them is difficult and leads to artifacts in the panorama synthesis. iii)~\textbf{Anchored}: We duplicate the input image to the backside, then generate the sky and ground, remove the backside, and then rotate the camera around~(\cref{fig:method-panos}, right). By anchoring the synthesis with global context, we are able to generate coherent equirectangular images.
For all heuristics, we need to further specify the resolution and field of view of the rendered perspective images, and the total number of generated views. The resolution is given by the resolution of the inpainting model. In our case, these are square images with 1024 pixels side length. For the middle region of the panorama images, we render 8 images with an 85-degree field of view. A large field of view ensures enough context, but the images also become increasingly distorted. For top and bottom regions, we use 4 images each, with a 120-degree field of view. We provide more qualitative results in \cref{sec:appAblations} of the supplementary document.

\boldparagraph{Prompt Generation.}
For zero-shot learning, the T2I model relies strongly on the given prompt to understand what it should inpaint. The most straightforward idea is to generate a caption from the input image. We use Florence-2~\cite{xiao2023florence} as off-the-shelf captioning model. However, using a description of the input image as prompt is insufficient, since the model fills all areas with duplications of the input image and fails to synthesize a reasonable spatial layout, see~\cref{fig:method-prompts} (left). Using a coarser description of the input image can help to remove duplications, but we often observe that sky and ground duplicate the scene, see~\cref{fig:method-prompts} (right). We hence resort to a vision-language model, Llama 3.2 Vision, and ask it to generate three prompts: i) A coarse description of the scene atmosphere, but ignoring central elements like objects and people. ii) A prompt for the sky or ceiling, depending on whether it is an indoor or outdoor scene. iii) A prompt for the ground or floor. Note that the model infers whether the scene is indoors or outdoors by itself, and we do not provide this information. \cref{fig:method-panos} (Anchored) shows the generated panorama image with directional prompts.

\boldparagraph{Refinement.}
To further improve the image quality, we found it beneficial to run a partial denoising process on the outpainted image. We use a standard text-to-image diffusion model and denoise using the last 30\% of the time steps. For a smooth transition, we create a soft mask by blurring the inpainting mask and use it to blend in the refined image.

\subsection{Point Cloud-Conditioned Inpainting}
The generated panorama image largely determines the content of the 3D scene. However, it only supports camera rotation and not translation. To make the 3D scene navigable on a VR headset, we need to lift it into three-dimensional space and fill in occluded areas.    

\methodLift
\boldparagraph{Panorama to Point Cloud.} 
To view the generated scenes on a VR device, the scale of the scenes should be approximately metric. We therefore consider Metric3Dv2~\cite{hu2024metric3d} as it is a state-of-the-art metric depth estimator. We render images from the generated panorama and predict their depth maps. The images are chosen to have overlapping regions, so that the predicted depths can be aligned and smoothly stitched together. However, we observe that even after filtering out low-confidence predictions, the predicted depth often produces distorted point clouds and places points too close to the camera, see \cref{fig:methodLift} for an example. In our setting, we find MoGE~\cite{wang2024moge} to be more robust, presumably due to its affine-invariant properties.
As MoGE's depth prediction is not metric, we align MoGE's depth $\mathbf{d}_{\text{MoGE}}$ with Metric3DV2's depth $\mathbf{d}_{\text{Metric3D}}$ by calculating a scaling factor $s_{\text{metric}}$ as follows:
\begin{eqnarray}
    s_{\text{metric}} &=& \frac{Q(0.8, \mathbf{d}_{\text{Metric3D}}) - Q(0.2, \mathbf{d}_{\text{Metric3D}})}{Q(0.8, \mathbf{d}_{\text{MoGE}}) - Q(0.2, \mathbf{d}_{\text{MoGE}})}\\
    \mathbf{d}_{\text{MoGE}}^\text{metric} &=& s_{\text{metric}} *\mathbf{d}_{\text{MoGE}}\,,
\end{eqnarray}
where $Q(p, \mathbf{x})$ returns the $p$th quantile of vector $\mathbf x$ 
We use quantiles for a more robust scale estimation. 
We observe that Metric3Dv2 often underestimates the scale of cartoonish-looking scenes. To counteract this, we additionally ensure that the average distance of the origin to the ground is at least 1.5m, where we consider all points with negative $z$ coordinate as part of the ground. 

\boldparagraph{Inpainting Occluded Areas.} 
When rendering a point cloud from a camera pose with translation, occlusions lead to empty areas in the novel views. We argue that filling these areas can be addressed as another inpainting task. Initial experiments reveal that off-the-shelf inpainting models struggle with the fragmented structures present in the rendered masks, as the training data typically consists of one continuous mask. Therefore, we fine-tune the inpainting model specifically on masks derived from point clouds.
We explore two strategies for generating training data for the model, both leveraging on-the-fly camera pose and point cloud estimation from CUT3R~\cite{wang2025cuter}. The first strategy involves constructing a point cloud from the input image, warping it to the novel view, and using the resulting warped image and mask as a condition for the model. In this case, the diffusion loss is applied to the novel view. Although CUT3R generally provides accurate predictions, they are not without errors. Warping inaccuracies from imprecise point clouds can lead to poor conditioning signals. In practice, we observe that with imperfect conditioning, the inpainting model struggles to adhere to the condition, as it cannot discern when the condition is accurate and when it should be disregarded. To overcome this, we revisit the approach proposed in~\cite{xiang2023ivid}. Instead of merely warping images to the novel view, we subsequently warp them back to the initial view. This forward-backward warping strategy, due to self-occlusions, produces similar masks on the input image. As the warped points are inherently correct, the conditioning signal for the model is also accurate, allowing the model to reliably adhere to the condition. We demonstrate this in~\cref{tab:pcdInpainting}. Our inpainting model is a combination of a T2I diffusion backbone and a ControlNet~\cite{zhang2023controlnet}. We fine-tune the model for only 5k iterations without any modifications to the architecture.\\
With the point cloud-conditioned inpainting model in place, the next step involves selecting appropriate camera poses to enhance the 3D scene. We construct a grid of camera poses that incorporate both rotation and translation. Let the origin be located at the center of a 2-meter cube. Cameras are positioned at the center of each of the six faces and at the eight corners of the cube, resulting in a total of 14 camera translation vectors. For each translation, we apply 14 distinct camera rotations. Six of these rotations align with the principal axes, directing the camera forward, backward, left, right, upward, and downward. The remaining eight rotations involve looking forward, backward, left, and right, each with a positive and negative roll of 45 degrees.

\panos
\subsection{3D Reconstruction}
\label{sec:3d_recon}
The 3D scenes are reconstructed using the images from both panorama synthesis and point cloud-conditioned inpainting. These images maintain the same resolution as the inpainting model, specifically $1024\times 1024$ pixels.
For the 3D representation, we select 3D Gaussian Splats~\cite{kerbl3Dgaussians} due to their high fidelity and fast rendering capabilities, specifically utilizing the Splatfacto implementation from NerfStudio~\cite{nerfstudio}.
We initialize the splats from the point cloud we obtain by lifting the panorama to 3D.
This already provides a very accurate, high-resolution initialization for the model, meaning that we can considerably shorten the standard Splatfacto training schedule to 5k steps, and disable periodic opacity reset.
Given that the point cloud-conditioned inpainting model may not always perfectly preserve the warped points, we restrict the use of these images to the inpainted regions for 3D reconstruction.
Conversely, for the generated images from panorama synthesis, we use the full image, except for the backside regions where the input image was initially placed as an anchor.

\boldparagraph{Trainable Image Distortion.} In order to account for small, local inconsistencies in the generated multi-view images, we augment Splatfacto with a trainable distortion model.
In particular, given a (pinhole) image $I$ rendered by GS, we resample it into a distorted image $\hat{I}$ according to
\begin{equation}
\label{eq:grid_distortion}
\hat{I}(\mathbf{p}) = \text{bilinear}(I; \mathbf{p} + f(\mathbf{p}, \mathbf{c}_I;\theta))\,,    
\end{equation}
where $\text{bilinear}(I; \mathbf{p})$ denotes bilinear interpolation of $I$ at normalized pixel coordinates $\mathbf{p}=(u, v)$, and use $\hat{I}$ instead of $I$ in the photometric losses during GS training.
The function $f(\mathbf{p}, \mathbf{c}_I;\theta)$ outputs an offset with respect to $\mathbf{p}$, given an image-specific embedding vector $\mathbf{c}_I$ and parameters $\theta$.
All image embeddings $\mathbf{c}_I$ and $\theta$ are optimized together with the 3D representation parameters as part of the standard GS training process.
In practice, we implement $f$ as a tiny MLP, and compute its values only on a low-resolution grid, bilinearly upsampling the result to the full resolution of $I$ before applying \eqnref{eq:grid_distortion}.
Please refer to \secref{sec:app3D} in the supplementary document for details.

\section{Experiments}
\panorama
\boldparagraph{Datasets.} We evaluate our recipe on both real photos and images produced by image generation models.
For the latter, we use the same input images as World Labs~\cite{worldlabs_blog} to facilitate qualitative comparisons.
For real-world images, we use the \emph{Advanced} collection from Tanks and Temples~\cite{Knapitsch2017SIGGRAPH} and select one image per scene.
Images are chosen avoiding people and non-descriptive close-up captures.
The list of filenames is provided in \cref{sec:appdatasets} of the supplementary document.
Our approach for panorama generation is training-free and hence does not require training data.
For point cloud-conditioned inpainting, we train ControlNet on DL3DV-10K~\cite{Ling2024DL3DV} and evaluate it on ScanNet++~\cite{Yeshwanth2023Scannetpp} as it contains ground-truth camera poses and depth.

\boldparagraph{Metrics.}
Since our problem setting is highly ambiguous and no ground-truth data is available for comparison, we focus our quantitative evaluation on measuring how well the generated environment aligns with the appearance of the input image, as well as on a number of image quality metrics, following~\cite{zhou2024dreamscene360}.
CLIP-I~\cite{hessel2021clipscore} measures the similarity between the CLIP image embeddings of novel images rendered from the synthetic scene and the input image.
NIQE~\cite{Mittal2013niqe}, BRISQUE~\cite{Mittal2012brisque}, and Q-Align~\cite{Wu2024qalign} are non-reference image quality assessment metrics.
As our goal is to create high-quality 3D worlds, assessing the image quality of the rendered 3D representation is a good proxy for the scene quality, as inconsistencies and reconstruction artifacts likely show up in the rendered images.

\boldparagraph{Implementation Details.} 
We use a transformer-based T2I inpainting diffusion model. Specifically, the model uses a ControlNet~\cite{zhang2023controlnet} to digest a masked input image in addition to a text prompt. Due to legal constraints, we use a proprietary model, but, since our recipe is nonspecific to the architecture of the inpainting model, publicly-available models could be adopted as well. 

\subsection{Panorama Generation}
\label{sec:panorama_generation}
We evaluate our training-free strategy against two publicly-available state-of-the-art methods: MVDiffusion~\cite{tang2023MVDiffusion} and Diffusion360~\cite{Feng2023Diffusion360}. The same text prompts are used across all models for a fair comparison. Qualitative results are presented in \cref{fig:panos}. Notably, MVDiffusion lacks support for synthesizing sky and ground of the panorama images, while Diffusion360 is prone to generating overly saturated textures and large patches of uniform color.
For quantitative evaluation, we render six images from each panorama, evenly distributed to cover a full 360-degree rotation around the z-axis, with a field of view set at 60 degrees. Due to MVDiffusion's limitation in handling upward and downward rotations, these viewing directions are excluded from the evaluation.
The results, as shown in \cref{tab:panorama}, demonstrate that our pipeline not only achieves the highest image fidelity but also aligns the panorama best \wrt the input image.

\subsection{Point Cloud-Conditioned Inpainting}
\inpainting
We evaluate two strategies for generating training data to fine-tune the inpainting model on rendered point clouds: forward warping with diffusion loss applied to the novel view, and forward-backward warping with diffusion loss applied to the masked input image. \cref{tab:pcdInpainting} shows that both methods are comparable \wrt image quality. However, forward-backward warping significantly enhances PSNR, suggesting that the model more effectively adheres to the input condition. These results corroborate our hypothesis that the quality of the condition is crucial for a good performance of the point cloud-conditional inpainting model. 

\subsection{3D Worlds}
\label{sec:3d_worlds}
\worlds
\gseval
Ultimately, our objective is to generate high-fidelity 3D worlds. We compare our pipeline with the best publicly-available baseline models: WonderJourney~\cite{yu2023wonderjourney} and DimensionX~\cite{sun2024dimensionx}. Both models produce videos as outputs and do not inherently provide a 3D representation. To address this, we create two trajectories, each performing a 180 degree rotation on a circle starting from the input image and rotating left and right, respectively. The camera is looking inwards, \ie, at the center of the circle. We then extract metric camera poses from the generated images using CUT3R~\cite{wang2025cuter}. For reconstructing the 3D Gaussian Splats, we use the same strategy as in our pipeline. \\
Qualitative results for the rendered 3DGS are presented in \cref{fig:worlds}. We observe that videos generated by WonderJourney can be inconsistent, hindering accurate pose extraction and 3D reconstruction. Consequently, the resulting 3D representation often overfits to individual images and overall contains many artifacts. DimensionX is more consistent and can produce good results within a limited range around the input image. However, minor inconsistencies in the generated videos are amplified during 3D reconstruction, decreasing sharpness of the 3D scenes.
By decomposing 3D synthesis into point cloud generation and subsequent inpainting, our pipeline produces more consistent outcomes, resulting in the sharpest 3D scenes with the highest fidelity. 
We quantify performance by evaluating the image quality of renderings from the 3D scenes. Specifically, we generate images from three distinct circular trajectories. The first trajectory maintains zero roll and zero z-translation, rotating at a radius of 0.5m around the origin while looking towards the scene center. The other two trajectories incorporate a roll of $\pm$45 degrees and a z-translation of $\mp$0.5m. For each trajectory, we render eight views with a 60-degree field of view at a resolution of 1024$\times$1024 pixels.
The quantitative evaluation in~\cref{tab:gseval} corroborates that our pipeline consistently obtains the best highest fidelity results. We provide additional qualitative results from our pipeline in~\cref{fig:teaser} and~\cref{fig:worldsOurs}. \\
We further analyze three different modules in our pipeline. First, we compare the downstream performance of the point cloud-conditional inpainting model against a variant that augments the point cloud using ViewCrafter~\cite{Yu2024ViewCrafter}. ViewCrafter is a state-of-the-art point cloud-conditioned video model, that generates a video based on a reference image and its warped point cloud. We render reference images from the panorama image and discard all generated frames except for the last, since inconsistencies in the generated videos can create artifacts in the 3D representation. 
We observe that our simple ControlNet approach results in better downstream performance and that our learnable grid distortion further improves robustness and details of the 3D scenes. 

\section{Limitations And Conclusion}
This paper outlines a recipe for generating 3D worlds from a single input image.
We show how to decompose this complex task into simpler subproblems, and propose strategic approaches to each of them using off-the-shelf methods, with minimal additional training effort required.
Thereby, the resulting pipeline remains generalizable and benefits from existing powerful generative models.
One remaining key challenge relates to the size of the navigable area in our generated worlds, as the complexity of the point cloud-conditioned inpainting task increases significantly beyond a 2-meter range from the initial viewpoint.
Generating the backsides of occluded areas is also currently out of reach.
Finally, our pipeline does not yet support real-time scene synthesis due to the inherent computational complexity associated with running inference on large-scale diffusion models.
However, once the 3D Gaussian Splats (3DGS) representation is created, it can be displayed in real-time on a VR device.

\clearpage
{
    \small
    \bibliographystyle{ieeenat_fullname}
    \bibliography{main,further_references,bibliography}
}
\clearpage
\appendix
\section*{\Large\textbf{Appendix}}
\section{Datasets}
\label{sec:appdatasets}
From Tanks and Temples, we use images with the following names: Auditorium: 00007.png, Ballroom: 00001.png, Courtroom: 00001.png, Museum: 00015.png, Palace: 00019.png, Temple: 00001.png. The images were selected such that there are no humans in the images and such that they show a larger scene, instead of close-up captures.

\section{Implementation Details}
For the equirectangular to projective projection, we add a slight blur to avoid sharp masking edges. We found that the inpainting models can be highly sensitive to such edge-artifacts, and an overall higher inpainting quality when ensuring smooth transitions. 

\boldparagraph{DimensionX.} We use their  \href{https://github.com/wenqsun/DimensionX}{official implementation} for our experiments. Specifically, we generate two trajectories. Both trajectories start from the input image. One rotates left and the other one right, moving along a circular trajectory and facing inwards. We tuned the pipeline to the best of our abilities but could not extend it to perform a matching full rotation, since the generated sequences do not produce overlapping content when meeting opposite of the input image.

\boldparagraph{WonderJourney.}
We use the \href{https://github.com/KovenYu/WonderJourney}{official code release} for our experiments. Similarly to DimensionX, we found it best to generate two camera rotations on an inward-facing circle, left and right, starting from the input image. Since there is no metric scale, we tune the hyperparamters to the best of our ability to make the radius of the circle reasonable.\\
For both approaches, we extract metric poses after generating the trajectories using CUT3R~\cite{wang2025cuter} as it robustly works with long input sequences. We use the resulting images and poses with the same 3D reconstruction settings as for our approach.

\subsection{3D Reconstruction - implementation details}
\label{sec:app3D}

\boldparagraph{Gaussian Splatting Training.}
We train our Gaussian Splatting models using the Splatfacto implementation from NerfStudio~\cite{nerfstudio}.
Most training hyper-parameters are kept at their default values, matching the original GS implementation from~\cite{kerbl3Dgaussians}, with the following changes:
\begin{itemize}
    \item we reduce the training schedule from 30k to 5k iterations;
    \item we disable periodic opacity reset, and keep adaptive density control active from iteration 500 to iteration 2500;
    \item we set the degree of the spherical harmonic functions that model view-dependent colors to 1;
    \item we increase batch size from 1 to 2.
\end{itemize}

\boldparagraph{Trainable Distortion.}
As mentioned in \secref{sec:3d_recon} in the main document, the function $f(\mathbf{p}, \mathbf{c}_I;\theta)$ that models the point sampling offsets in our trainable distortion model, is implemented as a small MLP.
Specifically, we use three linear layers with 128 hidden dimensions and ReLU activations, except for the last one that uses tanh.
Before feeding them to the MLP, the input positions $\mathbf{p}$ are encoded into 32-dimensional harmonic embeddings akin to NeRF~\cite{mildenhall2020nerf}.
The per-image codes $\mathbf{c}_I$ have 32 channels,
and the grid at which we evaluate $f$ has a resolution of $128\times 128$. We provide qualitative results with and without adding grid distortion in~\cref{fig:distortion}.

\section{Qualitative Ablation Studies}
\label{sec:appAblations}
\appAblationsA
\appAblationsB
As our heuristic for panorama synthesis is difficult to quantify, we instead provide extensive qualitative results in~\cref{fig:appAblationsA} and~\cref{fig:appAblationsB}. Similar to the results from the main paper, Ad-hoc synthesis yields pleasantly looking, yet geometrically incorrect results. Sequential synthesis often generates reasonable panorama images, too. However, we observe in multiple instances that the floor does not match the scene well. The anchored approach overall results in the best panorama images. We further compare results for using the image caption as prompt, finding that this often simply repeats characteristics of the input image in all directions. While the coarse prompt from a vision-language model improves on the duplications, the sky and ground-specific prompts used in the anchored approach further improve the results qualitatively.
In~\cref{fig:distortion}, we qualitatively show the effect of adding a trainable image distortion. While both approaches generate good results overall, grid distortion preserves more fine-grained details, \eg, for the branches of the trees.

\boldparagraph{Discussion About WorldLabs.ai Results.}
WorldLabs.ai recently introduced a solution for the single image to 3D generation task and presented their results on \href{https://www.worldlabs.ai/blog}{their blog}. Their approach is neither publicly disclosed nor are their resulting models available for a direct, side-by-side comparison. We observe that our results look sharper and provide a better sense of continuity. We also observe that our generated back side views are better aligned with respect to the style of the input image.  

\distortion
\fail
\boldparagraph{Failure Cases.} For challenging input images, \eg artworks with a distinct style, we observe that the initial outpainting can fail to faithfully extend the image \wrt the style, see~\cref{fig:fail}. This results in visible border artifacts around the input image. Further, we observe that occasionally the spatial layout of the panorama can be imperfect, even when using our anchored heuristic. While all synthesis results in this work were generated using a single seed, such failure cases could also be addressed with re-sampling or user edits on the prompt.

\end{document}